\documentclass{article}

\usepackage{PRIMEarxiv}

\usepackage[utf8]{inputenc} % allow utf-8 input
\usepackage[T1]{fontenc}    % use 8-bit T1 fonts
\usepackage{hyperref}       % hyperlinks
\usepackage{url}            % simple URL typesetting
\usepackage{booktabs}       % professional-quality tables
\usepackage{amsfonts}       % blackboard math symbols
\usepackage{nicefrac}       % compact symbols for 1/2, etc.
\usepackage{microtype}      % microtypography
\usepackage{lipsum}
\usepackage{array}
\usepackage{listings}
\usepackage{enumitem}
\usepackage{ragged2e}
\usepackage{relsize}
\usepackage{geometry} 
\usepackage{setspace}
\usepackage{float}
\usepackage{placeins}
\usepackage{tabularx}
\usepackage{booktabs}
\usepackage{amsmath}
\usepackage{siunitx}
\usepackage{pdfpages}
\usepackage{mathtools}
\usepackage{subcaption}
\usepackage[ruled,vlined]{algorithm2e}
\usepackage{fancyhdr}       % header
\usepackage{graphicx}       % graphics
\graphicspath{{media/}}     % organize your images and other figures under media/ folder

\title{Queue up for Takeoff: A Transferable Deep Learning Framework for Flight Delay Prediction
}

\author{
  Nnamdi Daniel Aghanya, Ta Duong Vu \\
  Cranfield University \\
  Cranfield\\
  \texttt{\{nnamdi.aghanya, duong.vu\}@cranfield.ac.uk} \\
   \And
  Amaëlle Diop, Charlotte Deville, Nour Imane Kerroumi \\
  Cranfield University \\
  Cranfield \\
  \texttt{\{amaelle.diop, charlotte.deville, nourimane.kerroumi\}@cranfield.ac.uk} \\
  \And
  Dr. Irene Moulitsas, Dr. Jun Li, Dr. Desmond Bisandu \\
  Cranfield University \\
  Cranfield \\
  \texttt{\{i.moulitsas, jun.li, desmond.bisandu\}@cranfield.ac.uk} \\
}

\begin{document}
\maketitle

\begin{abstract}
Air travel facilitates the movement of people and goods globally. However, flight delays remain a significant issue, profoundly causing financial losses, operational inefficiencies, and passenger dissatisfaction. Delay prediction remains a critical challenge in the aviation industry; and thus, several approaches have been developed to model and predict flight time delays in recent decades. Therefore, in order to provide great passenger experiences, boost revenue, and reduce unnecessary revenue loss, flight delays most be precisely predicted and such models can be generalised across networks. In this paper, a novel approach was developed that combines Queue-Theory and simple attention model, referred to as the Queue-Theory SimAM (QT-SimAM) model to predict flight delays. To validate our proposed model, we used data from the US Bureau of Transportation of Statistics. Our experiment results show that the proposed QT-SimAM (Bidirectional) out-performed the existing methods with accuracy of 0.927, precision of 0.946, recall of 0.927 and F1 score of 0.932. For transferability, we tested our model using the EUROCONTROL dataset. Our experiment results show that the proposed QT-SimAM (Bidirectional) has accuracy of 0.826, precision of 0.794, recall of 0.826 and F1 score of 0.791. Ultimately, the paper outlines an end-to-end methodology for predicting flight delays. Our suggested method's effectiveness in forecasting flight delays with the fewest possible errors is clearly defined by all evaluation parameters. Additionally, passengers' anxiety can be successfully reduced by using our robust model's prediction result to obtain information about the delayed flight ahead of time from aviation decision systems across networks.
\end{abstract}

% keywords can be removed

\section{Introduction}

Flight delays remain a significant issue in the aviation industry, profoundly causing financial losses, operational inefficiencies, and passenger dissatisfaction~\cite{Keselova2019, Khan2021, Chauhan2023}. Predicting flight delays remains a very difficult challenge due to their inherent, mostly inevitable uncertainty arising from a host of factors, such as weather conditions, air-traffic congestion effects, technical problems, airspace capacity, and airline-level operational constraints~\cite{Etani2019, Wang2022a, Carvalho2020}. Flight delays can be defined as departures that leave more than 15 minutes after their scheduled time or aeroplane arrivals that land over 15 minutes late, a benchmark set by the United States (US.) Department of Transportation and widely adopted in the literature~\cite{Gui2020, Mamdouh2024, Sternberg2017, PinedaJaramillo2024, Mokhtarimousavi2023}.

Airline delays result in significant financial and operational losses, costing the global industry billions annually~\cite{Lee2023, Bliman2010, Anupkumar2023}. A delay in one flight often disrupts subsequent flights due to the interconnected nature of airline schedules, affecting aircraft utilisation and crew assignments~\cite{Erdem2024}. Airlines incur additional costs from increased fuel consumption, crew hours, and passenger compensation~\cite{IATA2013}. Furthermore, delays have an environmental impact due to increased carbon emissions from extended taxiing or holding patterns~\cite{Sher2021}.

Delay prediction has been a critical challenge for decades, with several approaches developed to model and predict flight time delays~\cite{Koopman1972, Khanmohammadi2014, Tu2008, Cetek2013}. Fortunately, the modern aviation ecosystem generates vast amounts of data~\cite{ATIO2002}, and researchers have increasingly turned to advanced computational approaches like Machine Learning (ML) and Deep Learning (DL) to harness it~\cite{Kim2016}. These predictive models offer the potential to accurately forecast flight delays, enabling proactive decision-making~\cite{Carvalho2020}. However, results can be limited by the absence of specific airline data~\cite{Dalmau2021}, and despite extensive research, mitigating flight delays remains a primary problem to be solved~\cite{Birolini2023}. A significant gap in existing research is the reliance on weather-delay features for forecasting. This affects the model's utility for most open datasets that lack such features (e.g., the EU dataset), rendering a universally applicable solution impractical and highlighting the need for a transferable approach.

In response, this research predicts flight delays by developing a model that combines the Queue-Theory and simple attention model, referred to as the Queue-Theory SimAM (QT-SimAM) model. This architecture innovatively modifies the SimAM attention mechanism by biasing its energy function with proxies for an aircraft's accumulated workload, calculated using principles from Queue Theory. This integration allows the model to adaptively increase attention to feature channels associated with congested flight chains, providing a more informed representation of how delays cascade from one flight leg to the next. By addressing some key limitations in existing methodologies, such as the lack of generalizability and limited understanding of delay propagation, the study offers a robust framework for more accurate and transferable flight delay forecasting. The developed models provide technical information for airlines, airports, and air traffic management, demonstrating the potential for a broader application in different aviation networks (transferability). In addition, the research highlights the broader implications of advanced predictive technologies, offering a path to improved operational efficiency, passenger experience, and environmental sustainability in the aviation industry.

The rest of the paper is organised as follows. Chapter 2 establishes the flight delay methodology based on the Queue-Theory SimAM (QT-SimAM) model. In Chapter 3, we present prediction results and discussions, while Chapter 4 presents conclusions with limitations. The Appendix Chapter summarises related literature on flight time delays.

\section{Methodology}
\label{sec:Methodology}

This chapter details the methodology employed to develop and evaluate an attention-based framework for predicting flight delays, with a focus on assessing model transferability between the United States (US) and the European Union (EU). The core of this framework is a novel integration of queue-theoretic principles with attention mechanisms.

\begin{figure}[h!]
\centering
\includegraphics[width=0.8\textwidth]{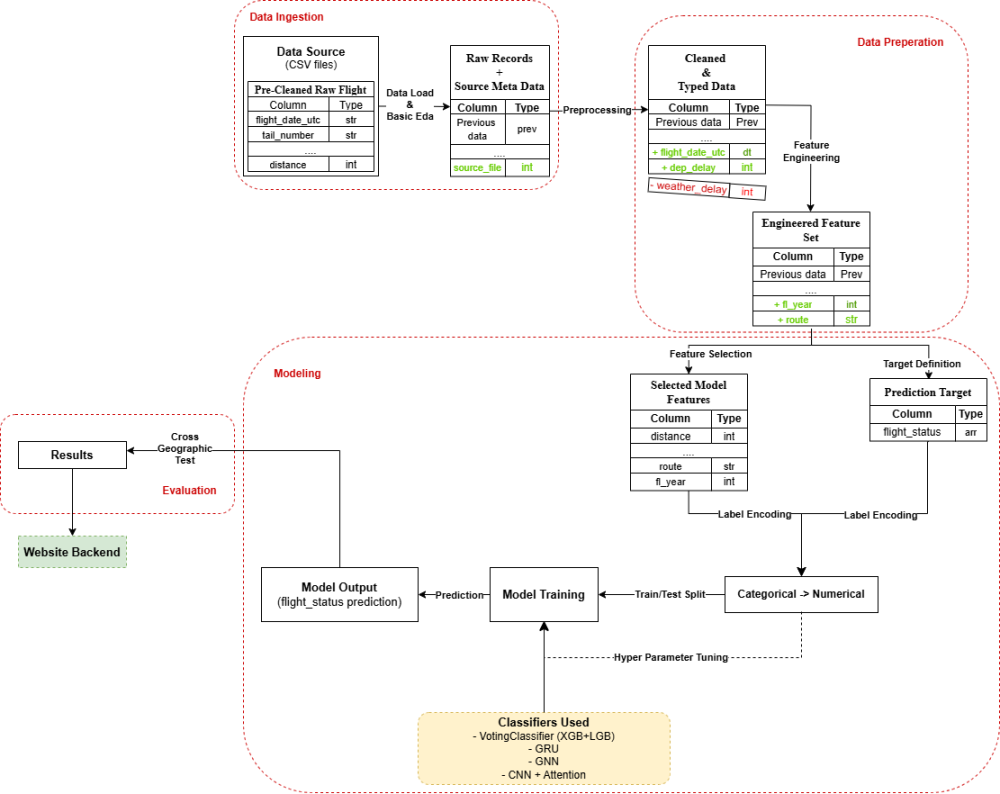} 
\caption{Flight Delay Prediction Pipeline Overview.}
\label{fig:MLPIPELINE}
\end{figure}

The subsequent sections describe the data acquisition and feature harmonisation steps, followed by the construction of flight chains for sequential modelling. The architectural details of the attention-based baseline models are then presented. The chapter culminates in the introduction of our proposed Queue-Theory Simple Attention Module (QT-SimAM) and its pairing with a Queue Mogrifier (QMogrifier) LSTM. An overview of the end-to-end workflow is provided in Figure~\ref{fig:MLPIPELINE}.

\subsection{Data Acquisition and Feature Harmonisation}
\label{sub:Data_Ingestion_Harmonisation}

This study utilises flight operations data from the US Bureau of Transportation Statistics and EUROCONTROL. Data from March, June, September, and December 2022 was selected to cover diverse operational and meteorological seasons, providing a robust baseline for evaluating model generalisation and transferability between the US and EU.

\vspace{0.5em}
\paragraph{Schema alignment}

The raw US and EU data files differ in scope and nomenclature. A notable asymmetry is a \texttt{weather\_delay} field in the US data with no counterpart in the European feed. A strictly comparable baseline, therefore, rests on the intersection of the two schemas as defined by Equation~\eqref{eq:intsc},
\begin{equation}
    \label{eq:intsc}
    \Sigma \;=\;\mathrm{Cols}\bigl(\widetilde{\mathcal{D}}^{\mathrm{US}}\bigr)\;\cap\;\mathrm{Cols}\bigl(\widetilde{\mathcal{D}}^{\mathrm{EU}}\bigr),
\end{equation}
which yields 38 shared features. To measure the incremental value of local features, an auxiliary set of US-only models is also trained with the \texttt{weather\_delay} feature reinstated.

Minor gaps were repaired by imputation: zeros for continuous variables and sentinel tokens for categoricals. No rows were dropped, avoiding region-specific sampling bias. The final harmonised datasets comprised approximately $N_{\text{US}} = 384,211$ (US) and $N_{\text{EU}} = 327,408$ (EU) rows.

\vspace{0.5em}
\paragraph{Feature catalogue}

In table~\ref{tab:corefeatures}, we list the key features comprising the harmonised set $\Sigma$. This selection ensures that all variables are available in both US and EU feeds, making the schema suitable for transfer experiments. Further, the set is consistent with the predictors used by Qu \emph{et al.}~\cite{qu2023flight}, allowing for direct comparisons to sequence-based models that have already proven effective.

\vspace{0.5em}
\paragraph{Implications for transferability}

A core objective is to evaluate model transferability, defined as the ability of a model trained on data from one region (e.g., the US) to predict outcomes accurately in another (e.g., the EU) using only the shared feature set. The primary transfer experiments are conducted using this restricted schema, providing a strict test of generalisation. The auxiliary US-only experiment with \texttt{weather\_delay} serves to quantify the performance gain from richer local data against the penalty incurred when such information is unavailable.

\begin{table}[htbp]
\caption{Unified feature catalogue after harmonised engineering}
\label{tab:corefeatures}
    \centering
    {
    \renewcommand{\arraystretch}{1.15}
    \setlength{\tabcolsep}{5pt}
    \footnotesize
    \begin{tabularx}{\textwidth}{
        >{\texttt\arraybackslash}X
        >{\raggedright\arraybackslash}X
        >{\centering\arraybackslash}X
    }
        \toprule
        \textbf{Feature name} & \textbf{Type} & \textbf{Role} \\
        \midrule
        distance & Numerical & Input \\
        flight\_date & Datetime & Input \\
        tail\_number & Categorical & Input \\
        airline & Categorical (label-encoded) & Input \\
        depart\_from\_iata & Categorical (label-encoded) & Input \\
        arrive\_at\_iata & Categorical (label-encoded) & Input \\
        scheduled\_departure\_utc & Numerical & Input \\
        actual\_departure\_utc & Numerical & Input \\
        departure\_delay\_minutes & Numerical & Input \\
        scheduled\_arrival\_utc & Numerical & Input \\
        scheduled\_estimated\_time & Numerical & Input \\
        arrival\_delay\_minutes & Numerical & \textbf{Output} \\
        \bottomrule
    \end{tabularx}
    }
\end{table}

\subsection{Flight-Chain Construction for Sequential Input}
\label{sec:flight_chain_construction}

Attention mechanisms are highly effective when supplied with temporally ordered data. We therefore replace isolated flight records with \emph{flight chains}: short, fixed-length sequences that mirror the daily schedule of an individual aircraft, adapting the method of Qu \emph{et al}.~\cite{qu2023flight}. Presenting consecutive flight legs in a single tensor allows the model to learn how delay on an early sector propagates through the remainder of the roster.

\paragraph{Formal definition}

\begin{figure}[h!]
\centering
\includegraphics[width=1\textwidth]{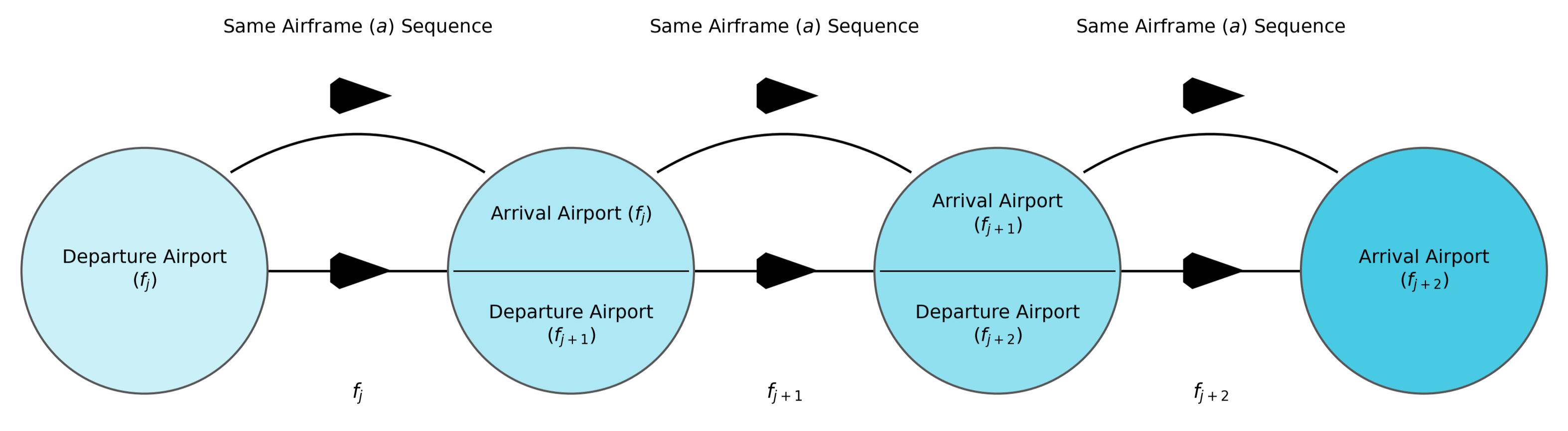}
\caption{Sequential Flight Input Representation (Flight Chain, $L=3$).}
\label{fig:flightchain}
\end{figure}

Let $a$ be an airframe and $t$ a date. The ordered set of sectors flown by $a$ on $t$ is
\begin{equation}\label{eq:sqf_new}
\mathcal{F}_{t}^{a}=\bigl(f_{1}^{a,t},\,f_{2}^{a,t},\dots,\,f_{\lvert\mathcal{F}_{t}^{a}\rvert}^{a,t}\bigr),
\end{equation}
where each leg $f_{k}^{a,t}$ has a feature vector $\mathbf{x}_{k}^{a,t}\in\mathbb{R}^{p}$. A sliding window of length $L=3$ with stride one extracts overlapping subsequences
\begin{equation}\label{eq:fchain_new}
\mathbf{c}_{j}^{a,t}=\bigl(f_{j}^{a,t},\,f_{j+1}^{a,t},\,f_{j+2}^{a,t}\bigr),\qquad
j=1,\dots,\lvert\mathcal{F}_{t}^{a}\rvert-L+1 .
\end{equation}
An aircraft completing at least three sectors thus yields $\lvert\mathcal{F}_{t}^{a}\rvert-2$ chains (see Figure~\ref{fig:flightchain}).

\vspace{0.5em}
\paragraph{Operational feasibility filter}

A chain is considered valid only if consecutive ground times respect
\begin{align}\label{eq:turnaround_new}
\texttt{Dep}_{j+1}&\ge\texttt{Arr}_{j}+\tau_{\min}, &
\texttt{Dep}_{j+2}&\ge\texttt{Arr}_{j+1}+\tau_{\min}, \notag\\
\Delta_{j,j+1}&\le\tau_{\max}, &
\Delta_{j+1,j+2}&\le\tau_{\max},
\end{align}
where $\tau_{\min}=15\,\text{min}$ and $\tau_{\max}=12\,\text{h}$. These thresholds mirror standard airline practice and ensure fair model comparisons.

\vspace{0.5em}
\paragraph{Tensor representation and target}
\label{sec:tensor}
For each feasible chain, the three feature vectors are stacked row-wise into a tensor $\mathbf{X}_{\mathbf{c}}\in\mathbb{R}^{3\times p}$.
\begin{equation}\label{eq:tensor_new}
\mathbf{X}_{\mathbf{c}}=
\begin{bmatrix}
\mathbf{x}_{j}^{\top}\\[2pt]
\mathbf{x}_{j+1}^{\top}\\[2pt]
\mathbf{x}_{j+2}^{\top}
\end{bmatrix}.
\end{equation}
The target label $y_{\mathbf{c}}\in\{0,1,2,3,4\}$ is derived from the arrival delay of the third leg, binned into five ordinal classes based on 15, 60, 120, and 240-minute cut-offs. This preserves hierarchical error information and supports class-balanced stratification. Table \ref{tab:delay-bins} summarises the ordinal target definition.

\begin{table}[htbp]
    \caption{Ordinal arrival-delay classes used for model training}
    \label{tab:delay-bins}
    \centering
    {
    \renewcommand{\arraystretch}{1.15}
    \setlength{\tabcolsep}{5pt}
    \footnotesize

    \begin{tabularx}{0.6\textwidth}{
        >{\centering\arraybackslash}X
        >{\raggedright\arraybackslash}X
    }
        \toprule
        \textbf{Ordinal label $y_{\mathbf{c}}$} & \textbf{Arrival-delay interval (min)} \\
        \midrule
        0 & $T \le 15$ \\
        \midrule
        1 & $15 < T \le 60$ \\
        \midrule
        2 & $60 < T \le 120$ \\
        \midrule
        3 & $120 < T \le 240$ \\
        \midrule
        4 & $T > 240$ \\
        \bottomrule
    \end{tabularx}
    }
\end{table}

\vspace{0.5em}
\paragraph{Chain-extraction algorithm and split}
The procedure outlined in Algorithm \ref{alg:chain-constructor_new} yielded approximately 280k chains from the US data and 15k from the EUROCONTROL data. Retaining identical preprocessing across regions is central to our transferability study. The US data was split into approximately 198k training, 42k validation, and 42k test examples. A robustness check using a stricter, non-overlapping split by tail-number-day confirmed equivalent accuracy, validating the use of the simpler window-based split.

\subsection{Attention models}
\label{sec:attention-models}
We implement two lightweight attention-based architectures as baselines, drawing on the CBAM and SimAM mechanisms~\cite{Woo2018, yang2021simam, qu2023flight}. Model hyperparameters such as channel widths and kernel sizes are kept constant across all experiments to ensure fair comparison.

\paragraph{CBAM–CNN feature extractor}
\label{sec:cbam-cnn}

Our first baseline is a one-dimensional Convolutional Neural Network (CNN) augmented with Convolutional Block Attention Modules (CBAM). The architecture consists of $K$ stages, each stacking a convolutional block and a CBAM gate~\cite{Woo2018}. A mini-batch $\mathbf{x}\in\mathbb{R}^{B\times S\times p}$ (batch size $B$, sequence length $S=3$, features $p$) is first permuted to $\mathbf{z}^{(0)}\in\mathbb{R}^{B\times p\times S}$ to treat features as channels. Each stage then computes:
\begin{equation}\label{eq:CBAM}
\mathbf{z}^{(\ell)}
  = \operatorname{CBAM}\Bigl(\,
      f_{\textsc{Conv}}^{(\ell)}(\mathbf{z}^{(\ell-1)})
    \Bigr),\quad
\ell=1\dots K.
\end{equation}
CBAM sequentially infers and applies channel and spatial attention maps ($\mathbf{M}_{\text{c}}$ and $\mathbf{M}_{\text{s}}$) to refine the feature representation:
\begin{subequations}
\label{eq:cbam_updated}
\begin{align}
\mathbf{M}_{\text{c}} &= \sigma\bigl(\operatorname{MLP}(\operatorname{AvgPool}(\mathbf{z})) + \operatorname{MLP}(\operatorname{MaxPool}(\mathbf{z}))\bigr) \\
\mathbf{z'} &= \mathbf{M}_{\text{c}} \odot \mathbf{z} \\
\mathbf{M}_{\text{s}} &= \sigma\Bigl(g_{k}\bigl[\operatorname{Mean}_{c}(\mathbf{z'}), \operatorname{Max}_{c}(\mathbf{z'})\bigr]\Bigr) \\
\text{CBAM}(\mathbf{z}) &= \mathbf{M}_{\text{s}} \odot \mathbf{z'}
\end{align}
\end{subequations}
After the final stage, global average pooling is applied across the sequence axis, followed by a fully connected layer to produce five-class logits. Empirical tuning led to $K=3$ stages with channel counts $\langle64,128,256\rangle$.

\paragraph{SimAM–CNN–LSTM Hybrid Model}
\label{sec:simam-cnn-lstm}

To better capture temporal dependencies, our second baseline couples a convolutional front-end with a recurrent back-end. The convolutional stack is enhanced with the parameter-free SimAM operator~\cite{yang2021simam}, which estimates saliency based on principles of neuronal inhibition. The architecture first passes the input through $K$ stages, each consisting of a convolution followed by SimAM:
\begin{equation}\label{eq:simAm}
\mathbf{z}^{(\ell)}=\operatorname{SimAM}\Bigl(f_{\textsc{Conv}}^{(\ell)}\bigl(\mathbf{z}^{(\ell-1)}\bigr)\Bigr),
\qquad\ell=1\dots K.
\end{equation}
The SimAM module assigns a saliency score to each neuron by calculating an energy function $y=\frac{(x-\mu_{c})^{2}}{4\bigl(\sigma^{2}_{c}+\lambda\bigr)}+0.5$, where $\mu_c$ and $\sigma_c^2$ are the channel mean and variance. The refined features are $\mathbf{z}_{\text{refined}}=\sigma(\mathbf{y})\odot\mathbf{z}$. The final feature map $\mathbf{z}^{(K)}$ is reshaped and forwarded to a single-layer LSTM, whose final hidden state is mapped to five-class logits.

\subsection{QTSimAM with QMogrifier LSTM}
\label{sec:qt-simam}
While the baseline models capture statistical patterns, they omit a key driver of cascading delays: the residual workload an aircraft carries between sectors. Drawing from queueing theory, an aircraft can be modelled as a single-server system whose workload evolves according to the Lindley recursion~\cite{Hernandez2015, Bae2020}. Our proposed model injects a proxy for this workload into both the attention mechanism (QT-SimAM) and the LSTM gating (QMogrifier LSTM).

\paragraph{Residual‑delay proxy}
Since an aircraft's complete history is often unavailable for real-time prediction, we compute queueing surrogates for each flight leg using its great-circle distance $d_t$ and airborne time $a_t$. These serve as proxies for service time and inter-arrival time, respectively. From these, we derive server utilisation $\rho$ and, using an $M/M/1$ approximation, the expected waiting time $W_q^{(t)}$ and queue length $L_q^{(t)}$:
\begin{equation}\label{eq:wst}
W_q^{(t)} = \frac{\rho}{1 - \rho + \varepsilon}\,E_S,
\qquad
L_q^{(t)} = \lambda\,W_q^{(t)}.
\end{equation}
These quantities are min-max normalised across the three-leg chain to produce $(W_n^{(t)},L_n^{(t)})\in^2$, providing a snapshot of the aircraft's congestion level.

\paragraph{Queue‑Theory SimAM}
We modify the SimAM energy function to direct attention towards legs burdened by residual delay. Let $\bar W_n$ and $\bar L_n$ be the chain-average normalised waiting time and queue length. The queue-aware energy becomes:
\begin{equation}\label{eq:qnrg}
e^{\star}=v(\mathbf z)+\bar W_n + 0.5\,\bar L_n + \varepsilon_\lambda,
\qquad
\varepsilon_\lambda=10^{-4},
\end{equation}
where $v(\mathbf z)$ is the original variance term. This biases the attention mask $\sigma(e^{\star})$ to preserve high-impact channels that signal a heavy workload, allowing the network to allocate capacity more effectively.

\paragraph{QMogrifier LSTM}
To make the recurrent head reactive to evolving queue dynamics, we adapt the LSTM cell. At each time-step $t$, the cell first mixes the previous hidden state $\mathbf h_{t-1}$ with the current queue proxies $W_n^{(t)}$ and $L_n^{(t)}$ to produce a gating mask $\mathbf{m}_t$.
\begin{equation}\label{eq:qmomod}
\mathbf m_t=\sigma\!\Bigl(
      [\,\mathbf h_{t-1};W_n^{(t)};L_n^{(t)}\,]^{\!\top}W_m+b_m
      \Bigr)
\end{equation}
This mask modulates the input feature vector, $\tilde{\mathbf x}_t=\mathbf m_t\odot\mathbf x_t$, before it enters a conventional LSTM cell, giving the recurrent layer direct sensitivity to workload fluctuations within the chain.

\paragraph{Back-propagation and Optimisation}
The complete network is trained end-to-end by minimising the Cross-Entropy loss using the Adam optimiser. All custom components—the residual-delay layer, QT-SimAM, and QMogrifier LSTM—are constructed from differentiable or sub-differentiable operations. This ensures that gradients can be computed for the entire parameter set $\boldsymbol{\Theta}$ via standard back-propagation, allowing the model to be tuned jointly. The parameter update follows the standard stochastic gradient descent rule:
\begin{equation}\label{eq:sgd_update_bp_section}
\boldsymbol{\Theta}_{t+1} = \boldsymbol{\Theta}_t - \eta_t \nabla_{\boldsymbol{\Theta}} \mathcal{L}_{B_t}(\boldsymbol{\Theta}_t)
\end{equation}
where $\mathcal{L}_{B_t}$ is the loss on a mini-batch $B_t$ and $\eta_t$ is the learning rate.

\section{Results}

\subsection{Environment Description}

The experimental setup consists of the Crescent HPC cluster based on CentOS 7 and comprises compute nodes with two Intel® Xeon® CPUs and NVIDIA® Tesla V100 GPUs. Each experimental run is allocated 96 CPU cores (six 16-core nodes) and can run up to two GPU-accelerated jobs in parallel.

\vspace{0.5em}
\subsection{Parameters Selection}
The hybrid model (QTSimAM-CNN-LSTM) evaluated in this study is configured with several hyperparameters that influence their training dynamics and predictive performance. General training parameters include the Adam optimiser \cite{Kingma2017}, a cross-entropy loss function, an initial learning rate of $10^{-4}$, and L2 weight decay (regularisation term $\lambda$) set to $10^{-5}$. The models were trained for a maximum of 50 epochs; this decision was informed by a careful review of related studies, particularly the work by Qu \emph{et al}. \cite{qu2023flight} indicated that similar model architectures often exhibit performance convergence around 40 epochs, stabilising loss and accuracy values. Opting for 50 epochs provides a sufficient margin for our models to reach a stable performance plateau. The training was carried out using the specified batch size of 32.

The architectural design incorporates a CNN component followed by an LSTM. The CNN typically consists of three convolutional layers with [64, 128, 256] output channels, respectively, utilising 3$\times$3 kernels and ReLU activation functions. The subsequent LSTM component is characterised by key hyperparameters such as its hidden state size, number of layers, bidirectionality, and dropout rate. While these hyperparameters offer avenues for optimisation through systematic tuning to tailor the model to specific dataset characteristics, the experiments presented in this paper utilise a consistent set of default values to ensure a fair baseline for model comparison. Specifically, for the LSTM, a hidden size of 256, 2 layers, and a dropout rate of 0.2 (applied within LSTM layers and before the final classification layer) were used.

The primary hyperparameters, including fixed training settings and default architectural choices used for this study, are summarised in Table \ref{tab:hyperparameters}. The table also indicates which parameters are generally considered tunable for further optimisation.

This standardised configuration allows for a focused evaluation of the model architectures themselves. Although hyperparameter tuning could yield further performance gains, establishing this baseline with default values is a crucial first step in understanding the inherent capabilities of the proposed models for flight delay prediction.

\begin{table}[htbp]
\centering
\caption{Experimental Hyperparameters for LSTM-based Models.}
\label{tab:hyperparameters}
\begin{tabular}{lcc}
\toprule
\textbf{Parameter Name} & \textbf{Default Value Used} & \textbf{Tunable} \\
\midrule
\multicolumn{3}{l}{\textit{General Training Parameters}} \\
Loss Function & Cross-Entropy Loss & No \\
Optimizer & Adam & No \\
Learning Rate & $1 \times 10^{-4}$ & Yes \\
Weight Decay ($\lambda$) & $1 \times 10^{-5}$ & Yes \\
Batch Size & 32 & Yes \\
Maximum Training Epochs & 50 & Yes \\
\midrule
\multicolumn{3}{l}{\textit{CNN Architecture Parameters}} \\
Kernel Size & 3$\times$3 & Yes \\
Output Channels & [64, 128, 256] & Yes \\
\midrule
\multicolumn{3}{l}{\textit{LSTM Architecture Parameters}} \\
Hidden Size & 256 & Yes \\
Number of Layers & 2 & Yes \\
Bidirectional & False & Yes \\
Dropout Rate & 0.2 & Yes \\
\bottomrule
\end{tabular}
\end{table}

\vspace{0.5em}
\subsection{Performance Metrics}
\label{sec:classification-metrics}

As mentioned earlier, this paper addresses the flight delay prediction task as a multiclass classification problem. Several standard performance metrics are employed to systematically evaluate the transferability and effectiveness of the proposed hybrid model. Primary metrics include overall accuracy, the confusion matrix, and per-class precision, recall, and F1 score. These metrics are derived from the confusion matrix, which provides a detailed breakdown of classification performance.

For a multiclass scenario with $C$ classes, the confusion matrix cross-tabulates the true classes against the predicted classes. For any given class $c \in C$:
\begin{itemize}
    \item \textit{True Positives (TP$_c$)}: The number of instances correctly predicted as belonging to class $c$.
    \item \textit{False Positives (FP$_c$)}: The number of instances incorrectly predicted as class $c$ (that is, they belong to another class $c' \neq c$).
    \item \textit{False Negatives (FN$_c$)}: The number of instances belonging to class $c$ but incorrectly predicted as another class $c' \neq c$.
    \item \textit{True Negatives (TN$_c$)}: The number of instances correctly predicted as not belonging to class $c$. (TN is less commonly used directly in multiclass P/R/F1 formulas for a specific class but is implicit in the sum of other classes).
\end{itemize}
\vspace{0.5em}
\subsubsection*{Accuracy.}
Accuracy represents the proportion of correctly classified instances in all classes to the total number of instances ($n$). It provides a general measure of the model's overall correctness and is calculated as shown in Equation~\eqref{eq:acc}:
\begin{equation}\label{eq:acc}
\text{Accuracy}= \frac{\sum_{i=1}^{n}\mathbf{1}\!\left[\hat{l}_i=l_i\right]}{n},
\end{equation}
where $\hat{l}_i$ is the predicted class for instance $i$, and $l_i$ is its ground-truth label. Although simple and intuitive, accuracy alone may not provide a complete picture of model performance in cases of significant class imbalance.
\vspace{0.5em}
\subsubsection*{Confusion Matrix.}
In this study, a $5 \times 5$ confusion matrix is used, corresponding to the five defined flight delay categories: Early/Slight Delay, Delayed, Significantly Delayed, Severely Delayed, Extremely Delayed (as detailed in \S\ref{sec:tensor}). True classes are matched against predicted classes. The diagonal elements of this matrix represent the correct classifications (TP$_c$ for each class $c$), while the off-diagonal elements highlight systematic misclassifications (FP$_c$ and FN$_c$).

\subsubsection*{Per-Class Metrics.}
From the components of the confusion matrix for each class $c$, the following class-specific scores are reported, as defined in Equation~\eqref{eq:class_metrics}:

\begin{equation}\label{eq:class_metrics}
\begin{aligned}
\text{Precision}_c &= \frac{\text{TP}_c}{\text{TP}_c+\text{FP}_c},\\
\text{Recall}_c &= \frac{\text{TP}_c}{\text{TP}_c+\text{FN}_c},\\
\text{F1-score}_c &= \frac{2 \times \text{Precision}_c \times \text{Recall}_c}
{\text{Precision}_c+\text{Recall}_c}.\\
\end{aligned}
\end{equation}

These metrics provide a nuanced understanding of performance for each specific delay category.
\begin{itemize}
    \item \textit{Precision$_c$} (P$_c$): For class $c$, this is the proportion of true positive predictions among all instances predicted as class $c$. It answers the question: "When the model predicts class $c$, how often is it correct?"
    \item \textit{Recall$_c$} (R$_c$): For class $c$, this is the proportion of true positive predictions among all actual instances of class $c$. It answers the question: "Of all true instances of class $c$, how many did the model correctly identify?"
    \item \textit{F1-score$_c$}: This is the harmonic mean of Precision$_c$ and Recall$_c$. The F1 score considers false positives (via precision) and false negatives (via recall). A high F1 score is achieved only when both precision and recall are high, making it a robust measure, especially when there is an uneven class distribution or when the cost of false positives and false negatives needs to be balanced.
\end{itemize}

Furthermore, macro-averaged and weighted-averaged scores for precision, recall, and F1 score are also computed from the per-class metrics to provide an overall estimate that is not biased by class imbalance. It is important to note that we chose the weighted-average scores because our dataset exhibits some notable class imbalance, and these scores more accurately reflect the model's overall performance by considering the proportion of each class in the actual data.

In summary, the combined use of accuracy, the detailed confusion matrix, and the per-class precision, recall, and F1 scores offer an informative and concise representation of the model's performance across the different flight delay categories.

\subsection{Analysis of Experimental Results}
This section analyses the experimental results, beginning with an in-region assessment of our proposed QTSIM models against contemporary deep learning benchmarks. This establishes the baseline performance within the primary operational domain before subsequent evaluations of model transferability.

\vspace{0.5em}
\subsubsection{In-Region Performance Evaluation}\label{sec:analysis}

This in-region evaluation involves training and testing the models exclusively on the US domestic flight dataset. Crucially, for this US-specific assessment, the input features for our QTSIM models include the \texttt{weather\_delay} feature, leveraging richer local information not available in the harmonised cross-regional schema. Table \ref{tab:performance_comparison} presents a comparative evaluation of our proposed QTSIM models against the CBAM-CondenseNet and SimAM-CNN-MLSTM benchmarks reported by Qu \emph{et al}. \cite{qu2023flight}. Our QTSIM with Bi-direction enabled (Bidir.) demonstrates superior performance across all key metrics, achieving the highest accuracy (0.93), precision (0.96), recall (0.93), and F1-score (0.93). This notably surpasses Qu \emph{et al.}'s leading SimAM-CNN-MLSTM model, which reported an accuracy of 0.9136 and an F1-score of 0.849, while our QTSIM (Bidir.) also matches its optimal loss value of 0.20. 

\begin{table}[htbp]
\caption{Performance comparison of CBAM-CondenseNet, SimAM-CNN-MLSTM, and QTSIM models}
\label{tab:performance_comparison}
{
\renewcommand{\arraystretch}{1.1}
\setlength{\tabcolsep}{5pt}
\scriptsize
\begin{tabularx}{\textwidth}{
    >{\centering\arraybackslash}p{2cm}
    >{\centering\arraybackslash}p{2.5cm}
    >{\centering\arraybackslash}p{2.5cm}
    >{\centering\arraybackslash}p{2.5cm}
    >{\centering\arraybackslash}p{2.5cm}
}
    \toprule
    \textbf{Metric} & \textbf{CBAM-CondenseNet} & \textbf{SimAM-CNN-MLSTM} & \textbf{QTSIM} & \textbf{QTSIM (Bidir.)} \\
    \midrule
    Loss value & 0.30 & 0.20 & 0.27 & 0.20 \\
    \midrule
    Accuracy & 0.898 & 0.914 & 0.855 & 0.927 \\
    \midrule
    Precision & 0.913 & 0.825 & 0.916 & 0.946 \\
    \midrule
    Recall & 0.892 & 0.874 & 0.855 & 0.927 \\
    \midrule
    F1 score & 0.904 & 0.849 & 0.870 & 0.932 \\
    \bottomrule
\end{tabularx}
}
\end{table}

\begin{figure}[h!]
\centering
\includegraphics[width=0.7\textwidth]{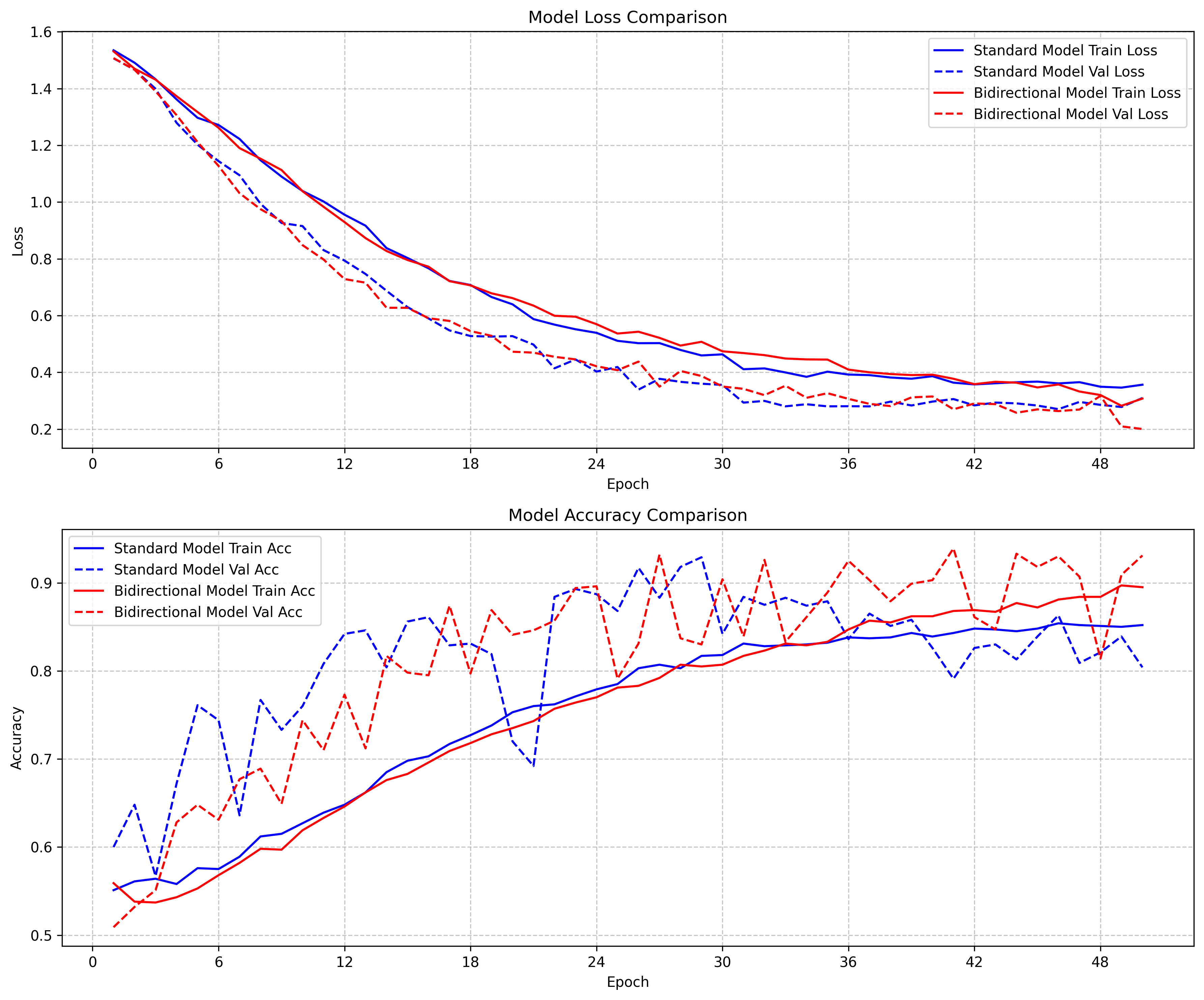}
\caption{Training loss and accuracy curves for QTSIM and QTSIM (Bidir.) models over 50 epochs on the US domestic dataset.}
\label{fig:loss_accuracy_curves}
\end{figure}

Figure \ref{fig:loss_accuracy_curves} shows the changes in loss and accuracy values for the QTSIM and QTSIM (Bidir.) models with the number of training epochs. Our models typically converge after approximately 38 epochs, demonstrating efficient learning; this convergence is slightly faster than the 40 epochs reported by Qu \emph{et al.} for their CBAM-CondenseNet, while their SimAM-CNN-MLSTM model converged more rapidly at 20 epochs.

\begin{table}[htbp]
\caption{Accuracy Comparison with Traditional and Deep Learning Models.}
\label{tab:comparison_with_traditional_models}
{
\renewcommand{\arraystretch}{1.1}
\setlength{\tabcolsep}{5pt}
\scriptsize
\begin{tabularx}{\textwidth}{
    >{\raggedright\arraybackslash}X
    >{\centering\arraybackslash}p{2.5cm}
}
    \toprule
    \textbf{Network Model} & \textbf{Accuracy (\%)} \\
    \midrule
    C4.5 Decision Tree \cite{qu2023flight} & 78.05 \\
    Support Vector Machine \cite{qu2023flight} & 80.00 \\
    ATD Bayesian Network \cite{qu2023flight} & 80.00 \\
    Artificial Neural Network \cite{qu2023flight} & 86.30 \\
    \midrule
    CBAM-CondenseNet \cite{qu2023flight} & 89.80 \\
    SimAM-CNN-MLSTM \cite{qu2023flight} & 91.36 \\
    \midrule
    \textbf{QTSIM (Bidir.)} & \textbf{92.76} \\
    \bottomrule
\end{tabularx}
}
\end{table}

To further contextualise the performance of our proposed approach, Table \ref{tab:comparison_with_traditional_models} extends this comparison to include a variety of traditional machine learning algorithms alongside the aforementioned deep learning architectures. These performance metrics for the benchmark models were taken from the work of Qu \emph{et al}.~\cite{qu2023flight}. The results show that in the case of the task of predicting flight delays, deep learning approaches outperform traditional methods. In particular, our QTSIM (Bidir.) model outperforms traditional algorithms by a large margin and surpasses previously reported deep learning models with a prediction accuracy of 93.00\%. These results provide a substantial performance baseline for our model architecture before we examine its ability to assess transferability.

\vspace{0.5em}
\subsubsection{Transferability Performance Evaluation}\label{sec:results_transfer}
Having established in-region benchmarks (\S\ref{sec:analysis}), we now assess model transferability: how well QTSIM models, trained on US data using only the harmonised feature set (excluding \texttt{weather\_delay}), generalise to the EU operational context. This tests the robustness of learnt delay patterns on unseen data from a different region.

Table \ref{tab:performance_harmonised_us} details the performance of the QTSIM models trained and tested on US data using only these harmonised features. As expected, excluding \texttt{weather\_delay} slightly reduces performance compared to Table \ref{tab:performance_comparison} (which included it). The QTSIM model achieved an accuracy of 0.913 (F1: 0.918), and the QTSIM (Bidir.) model an accuracy of 0.912 (F1: 0.917). These results serve as the direct baseline for evaluating transfer to the EU.

\begin{table}[htbp]
\caption{Performance of QTSIM models on US domestic dataset using only harmonised features.}
\label{tab:performance_harmonised_us}
{
\renewcommand{\arraystretch}{1.1}
\setlength{\tabcolsep}{5pt}
\scriptsize
\begin{tabularx}{\textwidth}{
    >{\centering\arraybackslash}X % Metric column
    >{\centering\arraybackslash}X % QTSIM (Harmonised)
    >{\centering\arraybackslash}X % QTSIM (Bidir., Harmonised)
}
    \toprule
    \textbf{Metric} & \textbf{QTSIM} & \textbf{QTSIM (Bidir.)} \\
    \midrule
    Loss value & 0.25 & 0.26 \\
    \midrule
    Accuracy & 0.913 & 0.912 \\
    \midrule
    Precision & 0.941 & 0.918 \\
    \midrule
    Recall & 0.913 & 0.912 \\
    \midrule
    F1 score & 0.918 & 0.917 \\
    \bottomrule
\end{tabularx}
}
\end{table}

Table~\ref{tab:transfer_performance_eu} contains the cross-regional evaluation results where models trained on US data were evaluated on previously unseen EU data. The models were evaluated with and without the US-specific \texttt{weather\_delay} feature. Using the harmonised feature set, QTSIM achieved an accuracy of 0.835 (F1: 0.815) on EU data, and QTSIM (bidir) achieved an accuracy of 0.826 (F1: 0.791). Models with the \texttt{weather\_delay} feature achieved varying transfer performance metrics (unidirectional: accuracy 0.819, F1 0.758; Bidirectional: accuracy 0.829, F1 0.827 on EU data). It is anticipated that we will experience some performance loss when transferring models to wildly different contexts, and these results suggest a good level of cross-regional generalisability. It should be noted that of the models trained on the harmonised feature set, the QTSIM model produced the highest transfer F1 score, indicating some evidence that training without region-specific features will perform better in "strict" transfer scenarios.

\begin{table}[htbp]
\caption{Transferability performance: QTSIM models trained on US dataset and tested on EU dataset.}
\label{tab:transfer_performance_eu}
{
\renewcommand{\arraystretch}{1.1}
\setlength{\tabcolsep}{5pt}
\scriptsize
\begin{tabularx}{\textwidth}{
    >{\raggedright\arraybackslash}p{4.5cm} % Model Configuration
    >{\centering\arraybackslash}X % Accuracy
    >{\centering\arraybackslash}X % Accuracy
    >{\centering\arraybackslash}X % Precision
    >{\centering\arraybackslash}X % Recall
    >{\centering\arraybackslash}X % F1-score
}
    \toprule
    \textbf{Model Configuration} & \textbf{Weather} & \textbf{Accuracy} & \textbf{Precision} & \textbf{Recall} & \textbf{F1-score} \\
    \midrule
    QTSIM & Yes & 0.819 & 0.777 & 0.819 & 0.758 \\
    \midrule
    QTSIM (Bidir.)& Yes & 0.829 & 0.826 & 0.829 & 0.827 \\
    \midrule
    QTSIM & No &0.835 & 0.812 & 0.835 & 0.815 \\
    \midrule
    QTSIM (Bidir.) & No& 0.826 & 0.794 & 0.826 & 0.791 \\
    \bottomrule
\end{tabularx}
}
\end{table}

To summarise, we found a decrease in performance when comparing models associated with the US domain and EU domain, which was expected in cross-regional contexts. However, QTSIM models provided a meaningful ability to generalise the learnt associations. To further evaluate the transferability of our models, we tested three established models (GNN, GRU network, and Voting classifier) under the same transfer conditions as the US to EU studies to provide comparison benchmarks. 

Table \ref{tab:transfer_accuracy_comparison_benchmarks} shows the transfer accuracy of the best QTSIM setups compared to a GNN, GRU, and Voting classifier setup. The results show that advanced deep learning models, specifically with QTSIM leading with 83.5\% accuracy, outperform established machine learning baselines in a challenging transfer learning domain. Not only does this setup outperform the Voting Classifier (81.1\%), it also outperforms the GRU (79.4\% accuracy) and GNN (59.6\%) baselines, adding to the confidence in the QTSIM architecture for cross-regional flight delay classification.

\begin{table}[htbp]
\caption{Transferability Accuracy Comparison: US-Trained Models Tested on EU Dataset.}
\label{tab:transfer_accuracy_comparison_benchmarks}
\centering
\begin{tabular}{lc}
\toprule
\textbf{Model} & \textbf{Transfer Accuracy (\%)} \\
\midrule
Graph Neural Network (GNN) & 59.6 \\
Gated Recurrent Unit (GRU) & 79.4 \\
Voting Classifier & 81.1 \\
\midrule
\textbf{QTSIM} & \textbf{83.5} \\
\bottomrule
\end{tabular}
\end{table}

\section{Conclusion}
This paper provides a methodology for flight delay prediction that emphasises data harmonisation and model transferability based on flight operations data from multiple sources and sequential input representations. This paper proposes three deep learning models for flight delay prediction alongside original queue-aware attention. Detailed layers of preparation and model design considerations have been outlined, and the conclusions are as follows.
\begin{enumerate}
    \item A robust data acquisition and feature harmonisation pipeline was established based on the need for cross-regional analysis. This process yielded a unified schema from US BTS and EU EUROCONTROL data, enabling the creation of "flight chains" temporally ordered sequences of flights, which allow models to learn how delays propagate through an aircraft's daily schedule, crucial for assessing model transferability.
    \item According to the spatiotemporal characteristics of flight delay propagation and the need for more nuanced attention, a QTSimAM-QMogrifier LSTM network was proposed integrating the principles of queueing theory. The convolutional front extracts spatial patterns, which are then weighted by a novel QT-SimAM module that incorporates residual delay proxies (derived from flight distance and airborne time) to focus attention on congested flight segments. Subsequently, a QMogrifier LSTM processes these queue-aware features, further extracting temporal dependencies by modulating its gates with per-leg delay proxies, thereby effectively improving the model's sensitivity to workload dynamics in predicting delay propagation.
\end{enumerate}

In conclusion, this paper outlines an end-to-end methodology for predicting flight delay outcomes, comprising data harmonisation, sequential input construction methods, and developing sophisticated, attention-based deep learning models. The application of queueing theory to the attention mechanism significantly contributes to operational awareness in flight delay predictions. Moving forward, we now look forward to a rigorous empirical evaluation of these models, including considering the transferability performance of these models between the US and the EU. Future work will also incorporate other harmonisable dynamic factors, such as real-time weather information and air traffic control data, to improve predictive accuracy under different operational circumstances. Furthermore, research could focus on advanced methodologies to address the inherent class imbalance in delay data. Enhancing model interpretability through advanced eXplainable AI techniques could also be explored to provide deeper insights into the causal drivers of predicted delays, fostering greater trust and operational utility.

%Bibliography
\bibliographystyle{unsrt}  
\bibliography{references}  

\appendix
\section{Literature Review}

With the rapid growth of the global aviation industry, flight delay is often acknowledged as one of the most important performance indicators in the sector~\cite{Sternberg2017}. Flight delays have severe negative repercussions, resulting in annual economic losses exceeding billions of dollars globally and causing significant operational inefficiencies and customer dissatisfaction~\cite{Ball2010, Liu2019, Atkinson2016, Anderson2009}. According to the Bureau of Transportation Statistics, both weather and non-weather factors can influence flight time delays.

\paragraph{Weather features on flight delays}
Weather-related factors significantly influence flight delays. Studies have shown that adverse weather conditions decrease airport capacity and cause delays throughout nearly every step of operations~\cite{Lall2018}. Research consistently finds that delays are more severe in bad weather, with thunderstorms, in particular, causing a considerable, non-linear increase in flight delays~\cite{Coy2006, Sasse2003}. Specific conditions like fog, thunderstorms, and snowfall can increase the chance of delay by over 25\%~\cite{Pejovic2009}. In quantitative terms, adverse weather can add from 10 to 23 minutes to departure delays, with factors like poor visibility and high winds being primary contributors~\cite{Borsky2019}.

\paragraph{Non-weather features on flight delays}
On clear days, non-weather factors—such as airline operations, maintenance, and crew problems—are the primary "culprits" of delays~\cite{Borsky2019}. Research has identified several key non-weather variables that impact delays, including temporal factors like the day-of-week and time-of-day~\cite{Rebollo2014, PerezRodriguez2017}. Other critical elements are the flight's origin, departure and scheduled arrival times, the distance between airports, and the specific airlines involved~\cite{Khanmohammadi2014, PerezRodriguez2017}. Furthermore, operational conditions such as air traffic control actions, the degree of airport crowdedness, and delays from a previous flight leg are also major contributors~\cite{Yu2019}.

\subsection{Flight Delay Prediction Models}
While researchers have long used traditional statistical approaches to model flight delays~\cite{Thiagarajan2017, Tu2008}, recent advances in big data analytics using machine learning (ML) and deep learning (DL) have shown great promise in improving predictive capacity~\cite{Rebollo2014, Kim2016, Lin2019}. This section reviews three main categories of flight delay models: simulation, queuing, and data-driven models~\cite{Ribeiro2025}.

\subsubsection{Simulation models}
Simulation models use a bottom-up method, simulating how an aviation system would operate based on predetermined rules. Common tools include SIMMOD, Total Airspace and Airport Modeller (TAAM), and AirTOP. SIMMOD has been used to analyse the causes of en-route delays and the impact of uncertainty in ground operations on delays~\cite{Cetek2013, Lee2012, Simic2020}. TAAM, which uses a 3D framework, has been employed to simulate airfield capacity and optimise aircraft trajectories to resolve conflicts~\cite{Offerman2001, MunozHernandez2017}. AirTOP is a fast-time simulator used to assess the impact of extreme weather events and to optimise flight trajectories to mitigate congestion~\cite{Gunther2015, Li2016, Kreuz2016}.

In summary, simulation models are valuable for assessing the potential effects of infrastructure and operational changes. However, their drawbacks include high computational costs, which can limit the scope of analysis, and the difficulty and time required for calibration and validation.

\subsubsection{Analytical: queuing models}
Analytical models use mathematical expressions of queuing dynamics to estimate flight delays, offering a computationally efficient alternative to simulation that does not require extensive data for calibration.

\paragraph{Stochastic and Non-stationary queueing models}
Stochastic queuing models are often characterized by probability distributions. The simplest forms, stationary models, assume that demand rate fluctuations are insignificant, a belief that is often incorrect in air travel where schedules change throughout the day~\cite{Bauerle2007, Grunewald2016, Odoni1983}. To address this, non-stationary queuing models relax the steady-state assumption. The DELAYS algorithm, based on $M(t)/E_k(t)/s$ queuing systems, is one of the most widely used and accurate analytical techniques for predicting flight delays and their propagation~\cite{Ribeiro2025, Koopman1972, Kivestu1976, Malone1995, Pyrgiotis2013, Lovell2007}.

\paragraph{Poisson and Pre-scheduled Random Demand Models}
Nonhomogeneous Poisson processes, which accommodate varying arrival and service rates, have also been applied to predict departure and operational delays~\cite{Hebert1997, Shone2019}. To overcome the limitations of Poisson models in precisely modelling arrival streams, the Pre-scheduled Random Demand (PSRD) or Pre-scheduled Random Arrivals (PSRA) model was introduced. The PSRD model offers more precise predictions and has been used extensively to investigate flight delay propagation and landing delays~\cite{Lancia2020, Nikoleris2012, caccava}.

In conclusion, queuing models offer a valuable and efficient method for predicting flight time delays, requiring fewer specific inputs than simulation models.

\subsubsection{Data-driven models}
Data-driven models use a top-down approach, leveraging historical data to identify trends and predict flight delays empirically.

\paragraph{Traditional Statistical Methods}
\label{sub:Traditional Statistical Methods}
Statistical models have been widely used to assess delay causes. Researchers have employed methods like multivariate simultaneous regression~\cite{Nayak2011, Hansen2005}, Cox proportional hazards models~\cite{Wong2012}, multinomial logistic regression~\cite{AbdelAty2007}, and ordered probit models~\cite{Pejovic2009} to identify factors associated with delays, such as season, flight distance, and time of day. Other techniques include Multivariate Adaptive Regression Splines (MARS)~\cite{Xu2008} and hybrid regression and time series methodologies~\cite{Markovic2008}.

\paragraph{Machine Learning Models}
\label{sub:Machine Learning}
Machine learning models often outperform traditional statistical methods, as they can automatically engineer high-order interactions between variables~\cite{Mokhtarimousavi2020, Yu2014}. Researchers have effectively implemented a range of ML algorithms, including random forests, decision trees, support vector machines (SVM), and neural networks to predict delays~\cite{Gui2020, Khaksar2019, Choi2016}. For instance, a combination of Bayes, rule, neural network, and decision tree models achieved a prediction accuracy of 79.73\% on data from a Chinese hub airport~\cite{Zonglei2008}. Other studies have used explainable AI (xAI) techniques like SHAP to interpret delay factors~\cite{PinedaJaramillo2024}, fuzzy inference systems to forecast delays at major airports~\cite{Khanmohammadi2014}, and hybrid feature selection methods to identify influencing variables~\cite{Dai2024}.

\paragraph{Deep learning models}
Deep learning (DL) represents a natural progression from ML, offering superior performance, particularly for long-range temporal dependencies. DL models excel at distilling hierarchical representations from complex data, which is critical for disentangling the multifactorial causes of delays~\cite{Kim2016}. Recurrent architectures like Long Short-Term Memory (LSTM) and frameworks combining Convolutional Neural Networks (CNN) with LSTMs have been used successfully to predict delays by capturing temporal patterns~\cite{Li2022, Li2023, Ai2019}.

While effective, these recurrent models often cannot model the spatial interdependencies between airports, limiting their ability to capture cascading delay propagation. This gap is addressed by recent advances in Graph Neural Networks (GNNs), which explicitly encode airport systems as graphs. This spatio-temporal modeling capability has positioned GNNs as an evolution beyond sequence-centric approaches~\cite{Cai2022, Guo2020b}. GNN architectures such as the Multiscale Spatial-Temporal Adaptive GCN (MSTAGCN)~\cite{Cai2022}, Deep Graph-Embedded LSTM (DGLSTM)~\cite{Zeng2021}, Adaptive Airport Awareness GNN (AAGNN)~\cite{Cai2024}, and ATFSTNP~\cite{Zang2022} have all demonstrated marked improvements in predicting delay propagation dynamics within complex air traffic networks.

\paragraph{Attention Mechanism}
\label{Attention Mechanism}
The attention mechanism was first developed for neural machine translation to allow a model to dynamically focus on the most relevant parts of a source input~\cite{Bahdanau2014}. This concept has since evolved into powerful architectures like the self-attention-based Transformer~\cite{Vaswani2017} and the Graph Attention Network (GAT)~\cite{Velickovic2018}. In transportation networks, this flexibility is highly attractive because delays propagate in a nonlinear, time-varying manner that is difficult to capture with fixed models. By allocating learnable weights to different airports and time horizons, attention mechanisms both boost predictive accuracy and provide an interpretable view of what interactions drive disruptions.

Building on this, recent aviation studies have integrated attention into sophisticated spatio-temporal models. For example, some models embed attention in a combinatorial framework to accelerate convergence~\cite{Fang2019}, while others use it to predict multi-step hourly delay dynamics~\cite{Bao2021}. The Spatio-temporal Graph Dual-Attention Network (SGDAN) provides real-time departure delay forecasts~\cite{Guo2020b}, and other recent models use multi-attention blocks to capture cross-airport contagion patterns or fuse bidirectional LSTMs with an attention head to refine flight-level estimates~\cite{Zheng2024, Mamdouh2024}. Collectively, these works demonstrate that attention is now a cornerstone in state-of-the-art delay-propagation models, enabling them to handle high-dimensional traffic data and the stochastic nature of real-world operations.

\section{Algorithms}

\begin{algorithm}[h]
\RestyleAlgo{ruled}
\DontPrintSemicolon
\KwData{Harmonised table $\widetilde{\mathcal{D}}$, window length $L$, turnaround limits $\tau_{\min}$ and $\tau_{\max}$}
\KwResult{Tensor stack $\mathcal{C}$ and label vector $\mathbf{y}$}
Sort $\widetilde{\mathcal{D}}$ by \texttt{Tail\_Number} and scheduled departure\;
\ForEach{same-tail, same-day block $g$}{
  \For{$j=1$ \KwTo $\lvert g\rvert-L+1$}{
        Form chain $c=(f_{j},f_{j+1},f_{j+2})$\;
        \If{turnaround rules (\ref{eq:turnaround_new}) hold}{
            Stack features to $\mathbf{X}_{c}$, append to $\mathcal{C}$\;
            Compute ordinal label $y_{c}$ from $f_{j+2}$, append to $\mathbf{y}$\;
        }
  }
}
Stratify $\mathcal{C},\mathbf{y}$ into train : val : test = 70 : 15 : 15\;
\caption{Extraction of feasible flight chains}
\label{alg:chain-constructor_new}
\end{algorithm}

\begin{algorithm}[h]
\RestyleAlgo{ruled}
\DontPrintSemicolon
\KwIn{$\mathbf{x}\in\mathbb{R}^{B\times S\times p}$}
\SetKw{Return}{return}
$\mathbf{z}\gets\mathbf{x}^{\top}$\tcp*{permute to (B,\,C,\,S)}
\For{$\ell=1$ \KwTo $K$}{
  $\mathbf{z}\gets \operatorname{CBAM}\bigl(f_{\textsc{Conv}}^{(\ell)}(\mathbf{z})\bigr)$
}
$\mathbf{h}\gets\operatorname{GAP}(\mathbf{z})$\tcp*{(B,\,C)}
\Return~$\mathbf{W}\mathbf{h}+b$\;
\caption{Forward pass of CBAM–CNN}
\label{alg:cbam-forward}
\end{algorithm}

\begin{algorithm}[h]
\RestyleAlgo{ruled}
\DontPrintSemicolon
\KwIn{Input mini-batch $\mathbf{x}\in\mathbb{R}^{B\times S\times p}$}
\SetKw{Return}{return}
$\mathbf{z}\gets\operatorname{permute}(\mathbf{x}, 0, 2, 1)$ \tcp*{To to (B, p, S) for Conv1d}
\tcp{Pass through K Conv+SimAM stages}
\For{$\ell=1$ \KwTo $K$}{
  $\mathbf{z}_{\text{conv}}\gets f_{\textsc{Conv}}^{(\ell)}(\mathbf{z})$
  $\mathbf{z}\gets \operatorname{SimAM}(\mathbf{z}_{\text{conv}})$
}
$\mathbf{s}\gets\mathbf{z}^{\top}$ \tcp*{Reshape to (B, S, C) for LSTM}
$(\_, (\mathbf{h}_{\text{T}}, \_))\gets\text{LSTM}(\mathbf{s})$ 

\If{LSTM is bidirectional}{
    $\mathbf{h}_{\text{final}} \gets \text{Concatenate final forward \& backward hidden states}$
} \Else {
    $\mathbf{h}_{\text{final}} \gets \mathbf{h}_{\text{T}}[-1]$ \tcp*{Get last layer’s hidden state}
}
\Return~$\mathbf{W}\mathbf{h}_{\text{final}}+b$\;
\caption{Forward pass of the SimAM–CNN–LSTM Model}
\label{alg:simam-cnn-lstm-forward}
\end{algorithm}

\begin{algorithm}[h]
\RestyleAlgo{ruled}
\DontPrintSemicolon
\caption{ResidualDelayLayer}
\label{alg:residual-rev}
\KwData{mini‑batch $x\in\mathbb R^{B\times S\times F}$}
\KwResult{$W_n,L_n\in[0,1]^{B\times S\times1}$}
\SetKw{Return}{return}

\BlankLine
\tcp{Extract distance and airborne‑time columns}
$d \gets x[:,:,F-3]$\;
$a \gets x[:,:,F-5]$\;

\BlankLine
\tcp{Queue parameters}
$E_S   \gets k_s\,d + \varepsilon$\;
$\lambda\gets k_a/(a + \varepsilon)$\;
$\rho   \gets \min(\lambda\,E_S,\,0.99)$\;

\BlankLine
\tcp{$M/M/1$ workload surrogates}
$W_q \gets \rho\;E_S/(1-\rho+\varepsilon)$\;
$L_q \gets \lambda\,W_q$\;

\BlankLine
\tcp{Min–max normalisation across the chain}
$W_{\min},W_{\max}\gets\min\nolimits_t W_q,\;\max\nolimits_t W_q$\;
$L_{\min},L_{\max}\gets\min\nolimits_t L_q,\;\max\nolimits_t L_q$\;
$W_n\gets\displaystyle\frac{W_q-W_{\min}}{W_{\max}-W_{\min}+\varepsilon}$\;
$L_n\gets\displaystyle\frac{L_q-L_{\min}}{L_{\max}-L_{\min}+\varepsilon}$\;

\BlankLine
\Return $W_n.unsqueeze(-1),\;L_n.unsqueeze(-1)$\;
\end{algorithm}

\begin{algorithm}[h]
\RestyleAlgo{ruled}
\DontPrintSemicolon
\caption{Forward pass of the \textsc{QT‑SimAM–CNN–QM‑LSTM} model}
\label{alg:qtsim-pipeline}
\KwIn{%
  flight‑chain tensor $\mathbf X\in\mathbb R^{B\times S\times p}$;
  \texttt{use\_softmp} $\in\{\texttt{true},\texttt{false}\}$}
\KwOut{classification logits $\hat{\mathbf y}$ and (optionally) residual‑delay prediction $\hat\Delta_S$}
\SetKw{Return}{return} % Ensure return is defined if needed, like the first example

\tcp{Calculate queue proxies per leg}
$W_n,\,L_n \gets \textsc{ResidualDelayLayer}(\mathbf X)$ \tcp*{Output shape: $(B,S,1)$}

\tcp{Apply optional Soft Max-Plus front-end}
\lIf{\texttt{use\_softmp}}{
      $\boldsymbol\delta\gets\textsc{SoftMaxPlus}(W_n\odot L_n,\tau)$\;
      $\mathbf X \gets \text{concat}(\mathbf X,\boldsymbol\delta)$
}

\tcp{Pass through CNN stem with QT‑SimAM}
$\mathbf Z \gets \text{permute}(\mathbf X,0,2,1)$ \tcp*{Reshape to $(B,C_0,S)$}
\tcp{Apply $L_{\text{conv}}=3$ Conv+QT-SimAM stages}
\For{$\ell\gets1$ \KwTo $L_{\text{conv}}$}{
  $\mathbf Z \gets \textsc{ConvBlock}^{(\ell)}(\mathbf Z)$ 
  $d \gets \text{mean}_t\, W_n$ 
  $l \gets \text{mean}_t\, L_n$
  $\mathbf Z \gets \textsc{QT‑SimAM}(\mathbf Z,d,l)$
}

\tcp{Process sequence with QMogrifier LSTM head}
$\mathbf S \gets \text{permute}(\mathbf Z,0,2,1)$ \tcp*{Reshape to $(B,S,C_{out})$}
$\_,\mathbf h_{T} \gets \textsc{QMogrifierStack}(\mathbf S,W_n,L_n)$ \tcp*{Get final hidden state $(B,H)$}

\tcp{Compute outputs}
$\hat{\mathbf y}\gets W_{\text{cls}}\mathbf h_{T}+b_{\text{cls}}$ \tcp*{Classification logits}
$\hat\Delta_S\gets W_\Delta\mathbf h_{T}+b_\Delta$ \tcp*{Residual delay prediction}
\Return~$\hat{\mathbf y}$, $\hat\Delta_S$\;
\end{algorithm}

\end{document}